\documentclass[10pt,final,journal]{IEEEtran} 

\usepackage{amsmath,amsfonts}
\usepackage{algorithmic}
\usepackage{algorithm}
\usepackage{array}
\usepackage[caption=false,font=normalsize,labelfont=sf,textfont=sf]{subfig}
\usepackage{textcomp}
\usepackage{stfloats}
\usepackage{url}
\usepackage{verbatim}
\usepackage{graphicx}
\usepackage{cite}
\usepackage{amsmath}
\usepackage{multirow}
\usepackage{xcolor}
\usepackage{diagbox}
\usepackage{colortbl}
\usepackage{arydshln}
\usepackage{booktabs}

\usepackage{tabularx}
\usepackage{romannum}
\usepackage{colortbl}  % 颜色表格支持
\usepackage{todonotes}

\definecolor{lightred}{RGB}{255, 200, 200}  % 浅红色
\definecolor{lightblue}{RGB}{200, 220, 255} % 浅蓝色
\definecolor{mycolor}{rgb}{0.902,0.902,0.980}
\definecolor{darkgreen}{rgb}{0.2,0.8,0.25}

\usepackage{CJKutf8}

\usepackage[normalem]{ulem}  % 导言区
\begin{document}

\title{SSVIF: Self-Supervised Segmentation-Oriented Visible and Infrared Image Fusion}

\author{Zixian Zhao, Xingchen Zhang$^{*}$}

\author{
	Zixian Zhao, Xingchen Zhang$^*$,\IEEEmembership{~Member, IEEE} \thanks{%This paper was supported by a Royal Society Research Grant (No. RG\textbackslash{}R1\textbackslash{}251462).
Z. Zhao and X. Zhang are with the Fusion Intelligence Laboratory, Department of Computer Science, University of Exeter, EX4 4RN, United Kingdom. (Email: zz541@exeter.ac.uk, x.zhang12@exeter.ac.uk) \\
\newline	$^*$ Corresponding author: Xingchen Zhang 
}
}

        % <-this % stops a space

\markboth{Journal of \LaTeX\ Class Files,~Vol.~14, No.~8, August~2025}%
{Shell \MakeLowercase{\textit{et al.}}: A Sample Article Using IEEEtran.cls for IEEE Journals}


\maketitle

\begin{abstract}
Visible and infrared image fusion (VIF) has gained significant attention in recent years due to its wide application in tasks such as scene segmentation and object detection.~VIF methods can be broadly classified into traditional VIF methods and application-oriented VIF methods. Traditional methods focus solely on improving the quality of fused images, while application-oriented VIF methods additionally consider the performance of downstream tasks on fused images by introducing task-specific loss terms during training. However, compared to traditional methods, application-oriented VIF methods require datasets labeled for downstream tasks (e.g., semantic segmentation or object detection), making data acquisition labor-intensive and time-consuming.~To address this issue, we propose a self-supervised training framework for segmentation-oriented VIF methods (\textbf{SSVIF}).~Leveraging the consistency between feature-level fusion-based segmentation and pixel-level fusion-based segmentation, we introduce a novel self-supervised task—cross-segmentation consistency—that enables the fusion model to learn high-level semantic features without the supervision of segmentation labels.~Additionally, we design a two-stage training strategy and a dynamic weight adjustment method for effective joint learning within our self-supervised framework. Extensive experiments on public datasets demonstrate the effectiveness of our proposed SSVIF.~Remarkably, although trained only on unlabeled visible-infrared image pairs, our SSVIF outperforms traditional VIF methods and rivals supervised segmentation-oriented ones. Our code will be released upon acceptance.
\end{abstract}


\section{Introduction}
\IEEEPARstart{V}{isible} and infrared image fusion (VIF) aims to generate a fused image that contains more useful information by leveraging the complementary characteristics of the two modalities \cite{ma2019infrared,zhang2023visible,liu2024infrared}. Visible images typically provide texture and color, while infrared images capture thermal radiation and offer stable imaging under challenging conditions such as low light or fog. In such environments, visible cameras often fail to produce informative images, limiting their effectiveness in tasks like scene segmentation and object detection. By combining these complementary sources, VIF has attracted increasing attention \cite{zhao2024equivariant, liu2024task, wu2025every, cao2024conditional, zhang2024e2e,cao2024test}, and plays a crucial role in enhancing the performance of downstream applications such as autonomous driving and robot perception.

With the advancement of deep learning, deep learning-based VIF methods have gradually become mainstream methods \cite{wu2025every, zhang2024mrfs, zhang2024text, cheng2025one}. These methods can be broadly categorized into two types: traditional VIF methods and application-oriented VIF methods. Traditional VIF methods primarily focus on optimizing the visual quality of the fused image during training. In contrast, application-oriented VIF methods not only consider visual quality but also introduce task-specific loss functions into the training process, thereby generating fused images more suited to specific downstream tasks.

\begin{figure*}[htbp]
	\centering
    \includegraphics[width=\textwidth]{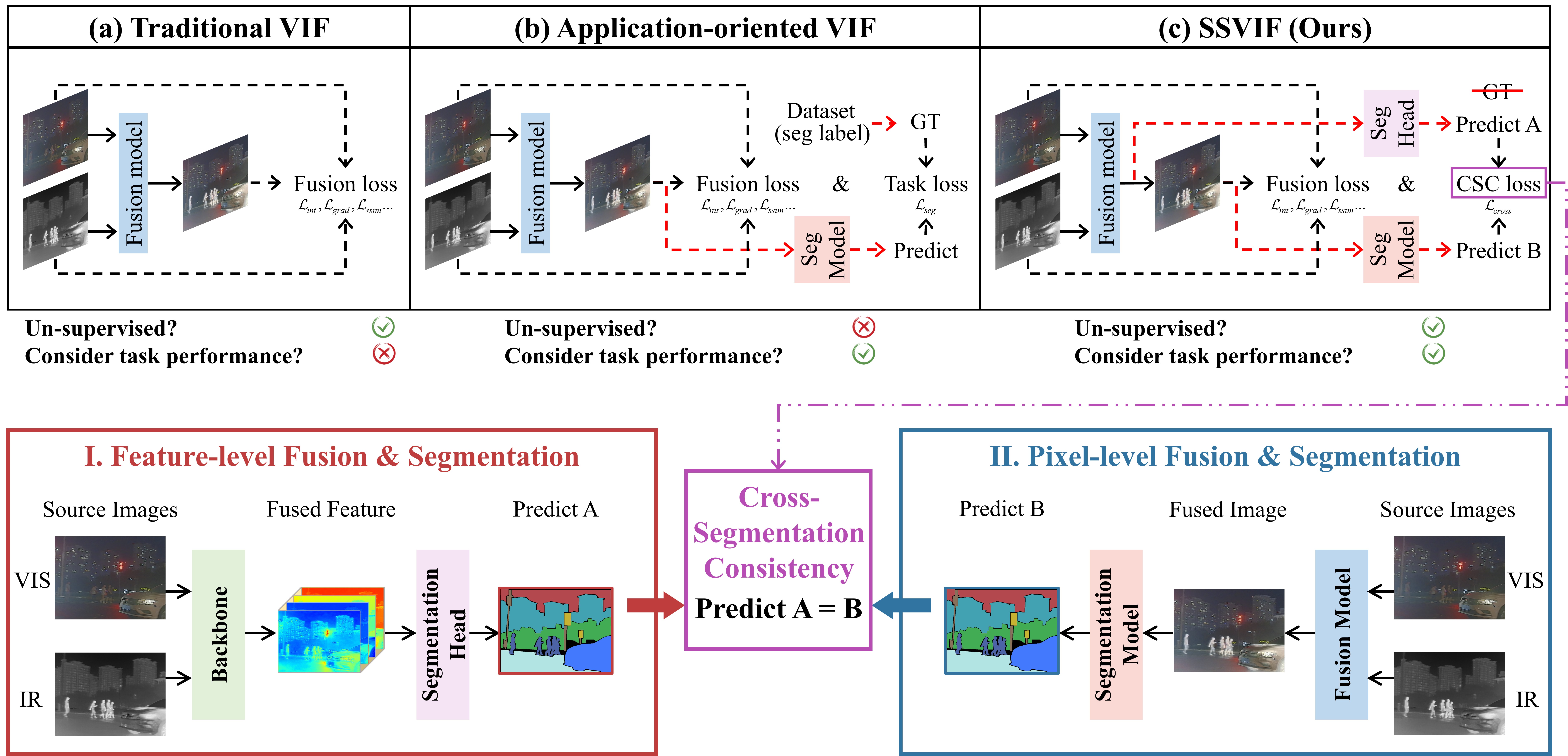}
	\caption{Overview of difference among (a) traditional VIF methods, (b) application-oriented VIF methods, and (c) the proposed SSVIF with a novel cross-segmentation consistency (CSC) task. CSC enables SSVIF to perform unsupervised learning while considering downstream task performance by performing segmentation from both fused features and fused images.}
	\label{fig:overview_first}
\end{figure*}

However, as shown in Fig. \ref{fig:overview_first}, unlike traditional VIF methods, which are generally unsupervised, existing application-oriented VIF methods are all supervised. That is, they require visible and infrared datasets annotated with task-specific labels (e.g., segmentation labels \cite{liu2023segmif, ha2017mfnet} or detection labels \cite{liu2022target}) during training. The presence of these labels enables the model to learn more task-relevant semantic features via task-specific loss functions. Nevertheless, the manual labeling process is time-consuming and labor-intensive, posing a major barrier to further research and application of these methods. To address this limitation, we propose \textbf{SSVIF}, a novel self-supervised training framework for segmentation-oriented VIF methods that does not require segmentation labels. At its core, SSVIF introduces a novel self-supervised task called \textbf{cross-segmentation consistency (CSC)} (bottom part of Fig.~\ref{fig:overview_first}).~Building on this self-supervised task, SSVIF adopts a two-branch structure and introduces an additional CSC loss to guide the fusion model to learn task-relevant semantic features in an unsupervised manner, thereby improving the segmentation performance of the fused images.

Moreover, since the fused images and the segmentation predictions are both of low quality in the early training phase, the CSC loss may fail to provide effective supervision for fusion model training. To address this issue, we design a \textbf{two-stage training strategy}~that decouples the early training of the fusion model and segmentation branches.~Additionally, we develop a gradient-and-descent-based weight adjustment (\textbf{GDWA}) method for joint training of the fusion and CSC tasks, which automatically balances the contributions of the fusion loss and CSC loss throughout training. This further improves both the visual quality of fused images and their performance in segmentation tasks.

In summary, the main contributions of this paper are as follows: \begin{itemize} 
    \item We propose a novel self-supervised training framework for segmentation-oriented VIF methods (SSVIF), which considers the segmentation performance of fused images during training, requiring no manual segmentation labels for supervision.
    
    \item We propose a novel self-supervised task, termed cross-segmentation consistency (CSC), for training fusion models. By utilizing the CSC loss, the fusion model is able to learn meaningful semantic features without relying on the supervision of any manual labels.
    
    \item We design a two-stage training strategy tailored to SSVIF and a novel dynamic weight adjustment method (GDWA) for joint training of the fusion and CSC tasks. Together, these approaches effectively balance the contributions of the fusion loss and the CSC loss, leading to improved overall model performance.

    \item Extensive experiments on public datasets demonstrate the effectiveness of the proposed SSVIF framework, including the CSC task and other key components. The results show that the fused images from SSVIF are better suited for downstream tasks like scene segmentation.
\end{itemize}

\section{Related work}
\subsection{Deep learning-based and Application-oriented VIF Methods}
In recent years, deep learning-based VIF methods have made significant progress, primarily focusing on enhancing fusion quality such as structural consistency and detail preservation. Early approaches, like Liu et al. \cite{liu2018infrared}, relied on CNNs to generate weight maps, while subsequent studies introduced architectures such as autoencoders \cite{li2019densefuse, li2021rfn}, GANs~\cite{ma2019fusiongan}, Transformers~\cite{ma2022swinfusion,rao2023tgfuse,zhao2024equivariant}, diffusion models~\cite{yue2023dif, zhang2024text}, and vision-language models~\cite{zhao2024image}. Alongside architectural advances, training strategies specifically designed for VIF have also been explored. For example, Li et al. \cite{li2021rfn} proposed a two-stage training strategy that separately trains an autoencoder and a residual fusion network, while Zhao et al. \cite{zhao2024equivariant} introduced a self-supervised loss exploiting geometric invariance. However, most existing methods focus only on visual quality, overlooking downstream utility. To address this, application-oriented VIF has emerged, aiming to optimize both image fusion and downstream task performance by incorporating task-specific losses into the training process~\cite{shopovska2019deep, tang2022image,zhang2023visible,liu2023segmif,zhang2024e2e,zhang2024mrfs}. For example, TarDAL~\cite{liu2022target} introduces a detection-based loss by running a detector on fused images, and MRFS~\cite{zhang2024mrfs} adopts a multi-task framework jointly optimized for segmentation and fusion. Recent studies~\cite{liu2023segmif, tang2022image, zhang2024mrfs, wu2025every} show that incorporating downstream task objectives can enhance both image quality and task performance of fused images. However, existing application-oriented VIF methods require manual labels for downstream tasks, which limits scalability due to high annotation costs.

\subsection{Self-supervised Learning}
Self-supervised learning (SSL) is an effective approach to address the high cost of label annotation. As a subset of unsupervised learning, SSL aims to learn discriminative features from unlabeled data~\cite{gui2024survey}, with the goal of narrowing the performance gap between unsupervised and supervised methods. The rapid development of SSL has gained significant attention in the research community~\cite{gui2024survey, liu2021self}, and it has been successfully applied to various computer vision tasks, including object tracking~\cite{bastani2021self}, image classification~\cite{assran2023self, gidaris2018unsupervised}, and image segmentation~\cite{fu2024bidcell, chaitanya2020contrastive, wang2020self}. Recently, semantic-consistency-based SSL has also been explored in other domains, such as weakly supervised semantic segmentation (via equivariant CAM consistency)~\cite{wang2020self} and single-view 3D reconstruction (via 2D–UV–3D part consistency)~\cite{li2020self}, demonstrating that semantic consistency can serve as a powerful supervisory signal. However, in the field of visible–infrared image fusion (VIF), such semantic-consistency-driven SSL solutions are still absent. In fact, most traditional unsupervised VIF methods can also be regarded as self-supervised approaches, since they rely on low-level reconstruction constraints (e.g., detail preservation, structural or spectral consistency) between fused and source images. Although several more recent self-supervised VIF methods have been proposed \cite{liu2024stfnet, li2024lrfe, zhang2022ssl}, they likewise focus on low-level consistency to enhance perceptual quality, but still overlook downstream task performance. In contrast, existing application-oriented VIF methods are either weakly supervised \cite{shopovska2019deep, wu2025every} or fully supervised \cite{liu2023segmif, tang2022image, zhang2024mrfs}, thus requiring costly annotations. This gap motivates the design of a self-supervised, application-oriented VIF framework. Our SSVIF addresses this by introducing cross-segmentation consistency (CSC) to enable fusion models to learn more semantic features during training without requiring segmentation labels.

\section{The Proposed Method}
\begin{figure*}[t]
	\centering
	\includegraphics[width=\textwidth]{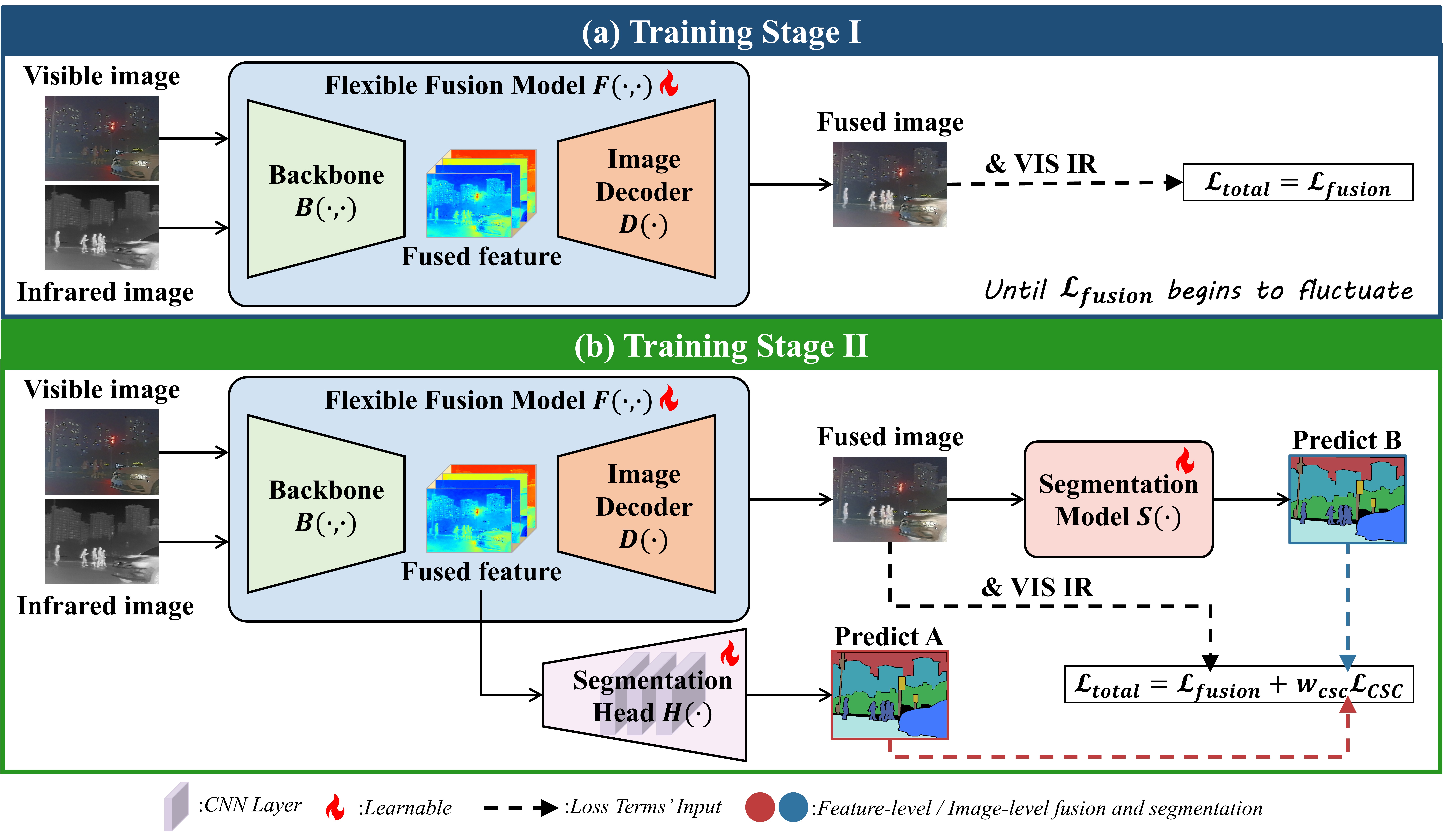}
	\caption{The workflow of proposed SSVIF training framework. During training, visible and infrared images are simultaneously fed into a flexible fusion model $F$ for feature extraction, fusion, and image reconstruction. Segmentation model and head are used to get CSC loss. After training, the fusion model inside the light blue box is used to generate fused images during inference.}
	\label{fig:fusion}
\end{figure*}

\subsection{Method Overview}
The goal of our work is to enable the fusion model to learn not only low-level features but also high-level semantic features in an self-supervised manner during training. To achieve this goal, we propose a novel self-supervised task for application-oriented VIF methods: cross-segmentation consistency (CSC). Specifically, as shown in Fig. \ref{fig:overview_first}, for the same pair of visible and infrared images, the segmentation predictions based on feature-level fusion should be consistent with those based on pixel-level fusion. This consistency exists because, in multimodal segmentation tasks, both the fused features and the fused images are ultimately used to produce accurate segmentation results, which are one-to-one aligned with the inputs. Therefore, when the inputs are the same, the segmentation predictions obtained from these two fusion pathways should remain consistent. Building on this idea, we propose a novel segmentation-oriented self-supervised VIF training framework, named SSVIF.

\subsection{Cross-segmentation Consistency}
\label{subsec:csc}
Specifically, let $I_{vis}\in \mathbb{R}^{3 \times H \times W}$ and $I_{ir}\in \mathbb{R}^{3 \times H \times W}$ denote visible and infrared images, respectively. For any fusion model $F(\cdot,\cdot)$ composed of a backbone $B(\cdot,\cdot)$ ~and an image decoder $D(\cdot)$, we introduce a trainable segmentation head $H(\cdot)$ and a trainable segmentation model $S(\cdot)$. Based on this setup, the segmentation predictions from the feature-level and pixel-level fusion pathways can be expressed as $\hat{p}^\text{A} = H(B(I_{ir},I_{vis}))$ and $ \hat{p}^\text{B}=S(F(I_{ir},I_{vis}))$, respectively. The objective of the cross-segmentation consistency task can thus be formulated as: 
\begin{equation}
\label{eq.CSC_target}
\hat{p}^\text{A}=\hat{p}^\text{B} \Leftrightarrow H(B(I_{ir},I_{vis})) = S(F(I_{ir},I_{vis})),
\end{equation}
where the fusion model is given by $F(I_{ir},I_{vis})=D(B(I_{ir},I_{vis}))$. Through this task design, the fusion model is encouraged to learn high-level semantic features without relying on the supervision of segmentation labels. Overall, the optimization problem of our SSVIF framework consists of two tasks: the fusion task and the CSC task, formulated as:

\[
\begin{cases}
\text{Task 1 (Fusion Task):} &  \displaystyle \min_{\omega_f} \; \mathcal{L}_{fusion}(I_{f}, I_{ir},I_{vis}), \\
\text{Task 2 (CSC Task):} &  \displaystyle \min_{\omega_f, \omega_h, \omega_s} \; \mathcal{L}_{csc}(\hat{p}^\text{A}, \hat{p}^\text{B}),
\end{cases}
\]
where $I_{f}=F(I_{ir},I_{vis})$ denotes the fused image, $\omega_f,\omega_h,\omega_s$ denote the learnable parameters of fusion mdoel, segmentation head, and segmentation model, respectively. The fusion loss $\mathcal{L}_{fusion}$ and CSC loss $\mathcal{L}_{csc}$ are detailed in Section \ref{sec:loss}.

\subsection{SSVIF Training Framework}
\label{sec:SSVIF}
\noindent\textbf{Dual Segmentation Branches.}~To construct the cross-segmentation consistency, we need to obtain segmentation predictions from both feature-level and pixel-level fusion.~To achieve this, as shown in the bottom part of Fig. \ref{fig:fusion}, we connect a segmentation head and a segmentation model to the fused features and the fused image, respectively. The segmentation head comprises two 3×3 CNN layers followed by a 1×1 CNN layer that outputs an $n$-class segmentation map. The segmentation model adopts SegFormer-B3~\cite{xie2021segformer}, which also produces an $n$-class segmentation map.~It is worth noting that, unlike previous weakly supervised VIF methods (e.g., SAGE \cite{wu2025every}) that rely on pre-trained segmentation models, both our segmentation head and segmentation network are trained entirely from scratch without using any pretrained weights. For more details about the setting of $n$, see Appendix \ref{ap.hyperparameter}.

\noindent\textbf{Two-stage Training Strategy.}
Achille et al.~\cite{achille2017critical} found that introducing low-quality or misleading data during the early phase of training can cause lasting damage to a model’s final performance. Inspired by this insight, we note that the dual segmentation branches, being trained from scratch, cannot provide reliable CSC supervision for the fusion model in the early training phase.~At the same time, the fusion model—also initialized from scratch—initially produces low-quality fused features and images, which in turn hinder the learning of the segmentation branches. To address these issues, we propose a two-stage training strategy for SSVIF, aiming to decouple the early training phases of the fusion model and the segmentation branches.

In Stage I, as shown in Fig. \ref{fig:fusion} (a), only the parameters of the fusion model are updated using the fusion loss $\mathcal{L}_{fusion}$, and the average fusion loss $\mathcal{L}_{fusion}^{j}$ over all steps within each epoch $j$ is recorded. When the recorded loss begins to increase, i.e., $\mathcal{L}_{fusion}^{j}>\mathcal{L}_{fusion}^{j-1}$, we consider the early training phase of the fusion model to be complete. At this point, SSVIF switches to Stage II of training.

In Stage II, as shown in Fig.~\ref{fig:fusion} (b), the segmentation head and segmentation model are incorporated into joint training with the fusion model. The segmentation predictions from both feature-level and pixel-level fusion are used to compute the cross-segmentation consistency loss $\mathcal{L}_{csc}$. The overall supervision in this stage combines $\mathcal{L}_{fusion}$ and $\mathcal{L}_{csc}$, jointly guiding the training of the fusion model and the two segmentation branches. This two-stage training strategy effectively enhances the stability and final performance of SSVIF, as further demonstrated by the ablation studies in Section \ref{sec:ablation}.

\noindent\textbf{Flexible Fusion Model.}
Most deep learning-based VIF methods consist of a backbone and an image decoder, where the backbone performs feature extraction and fusion to generate fused features, and the image decoder reconstructs the fused image from these features. Since SSVIF only requires access to fused features and fused images during training, it is not tied to any specific fusion model architecture and can be adopted as a general training framework to a wide range of fusion models.

\subsection{Loss function}
\label{sec:loss}
In SSVIF, the fusion loss $\mathcal{L}_{fusion}$ and the cross-segmentation consistency loss  $\mathcal{L}_{csc}$ are adopted to give low-level and high-level supervision during the training process, respectively. The total loss $\mathcal{L}_{total}$ for SSVIF can be calculated as:
\begin{equation}
\label{equ:total_loss}
\mathcal{L}_{total} =
\begin{cases}
\mathcal{L}_{fusion}, & \text{Stage I} \\
\mathcal{L}_{fusion} + \omega_{csc} \mathcal{L}_{csc}, & \text{Stage II}
\end{cases}
\end{equation}
where $\omega_{csc}$ is a dynamic weight for $\mathcal{L}_{csc}$, which is described in Section \ref{sec:GDWA}.

\noindent\textbf{Fusion Loss.}~We adopt commonly used fusion loss terms from existing VIF methods to construct the fusion loss, aiming to preserve pixel intensity, texture detail, structural information, and color distribution in the fused images. Specifically, the fusion loss includes the following components: intensity loss $\mathcal{L}_{int}= \frac{1}{HW}  \| I_f - \max(I_{ir}, I_{vis}) \|_1$, gradient loss $\mathcal{L}_{grad} = \frac{1}{HW}  \| |\nabla I_f| - \max( |\nabla I_{ir}|, |\nabla I_{vis}| ) \|_1$, structural similarity loss $\mathcal{L}_{ssim}= \mathbb{E}_{j \in \{ir, vis\}} \left( 1 -  ssim(I_f, I_j)  \right)$, and color-preserving loss $\mathcal{L}_{color} = \frac{1}{HW} \mathbb{E}_{c \in \{Cb, Cr\}} \| I_f^c - I_{vis}^c \|_1$.~Here, $\mathbb{E}$ is the expectation operator, and $\nabla$ is the Sobel gradient operator. $\left|\cdot\right|$ and $\left\|\cdot\right\|_1$ indicate the absolute value and $l_1$-norm operations, respectively. $ssim(\cdot)$ is the structural similarity index~\cite{wang2004image}.

In summary, our fusion loss $\mathcal{L}_{fusion}$ contains four loss terms and is calculated as:
\begin{equation}
\label{eq.fusion}
\mathcal{L}_{fusion} = \lambda_1 \mathcal{L}_{int} + \lambda_2 \mathcal{L}_{grad} + \lambda_3 \mathcal{L}_{ssim} + \lambda_4 \mathcal{L}_{color},
\end{equation}
where $\lambda_1,\lambda_2,\lambda_3,\lambda_4$ are hyper-parameters controlling the trade-off of each sub-loss term. For hyper-parameter settings in our practical experiments, see Appendix \ref{ap.hyperparameter}.

\noindent\textbf{Cross-segmentation Consistency Loss.}
To enforce consistency between the dual segmentation branches, we design a cross-segmentation consistency loss $\mathcal{L}_{csc}$ based on the idea introduced in Section \ref{subsec:csc}. Firstly, we obtain two segmentation predictions from the feature-level ($\hat{p}^\text{A}$) and pixel-level ($\hat{p}^\text{B}$) branches.~For each pixel, we compare the predicted class confidence scores from both branches, and select the higher-confidence prediction to construct pseudo label as follows:
\begin{equation}
\label{eq.pseudo_labels}
\tilde{p}_c(x) =
\begin{cases}
\arg\max_c \hat{p}_c^\text{A}(x), & \text{if } \max_c \hat{p}_c^\text{A}(x) > \max_c \hat{p}_c^\text{B}(x) \\
\arg\max_c \hat{p}_c^\text{B}(x), & \text{otherwise}
\end{cases}
\end{equation}
where $\tilde{p}_c(x)$ denotes the probability that pixel $x$ belongs to class $c$ in pseudo label $\tilde{p}$. $\hat{p}_c^\text{A}(x)$ and $\hat{p}_c^\text{B}(x)$ are similar probabilities from $\hat{p}^\text{A}$ and $\hat{p}^\text{B}$.

Then, the selected pseudo label $\tilde{p}$~is used for supervision via a hybrid loss $\mathcal{L}_{hyb}$, which combines cross-entropy $\mathcal{L}_{ce}$ and Dice loss  $\mathcal{L}_{dice}$ \cite{sudre2017generalised} as follows:
\begin{equation}
\label{eq.hybrid_loss}
\begin{cases}
\mathcal{L}_{ce} = - \sum_{x} \log \hat{p}_{p(x)}(x), \\[8pt]
\mathcal{L}_{dice} = 1 - \dfrac{1}{n} \sum_{c=1}^{n} \dfrac{2 \sum_x \hat{p}_c(x) \cdot p_c(x)}{\sum_x \hat{p}_c(x) + \sum_x p_c(x) + \epsilon}, \\[8pt]
\mathcal{L}_{hyb} = \mathcal{L}_{ce} + \mathcal{L}_{dice}
\end{cases}
\end{equation}
where $n$ is the number of segmentation classes introduced in Section \ref{sec:SSVIF} and $\epsilon$ is a smoothing term introduced to prevent division by zero. Finally, the total consistency loss $\mathcal{L}_{csc}$ is the average of the two hybrid losses computed from each branch's prediction as:
\begin{equation}
\label{eq.csc_loss}
\mathcal{L}_{csc} = \frac{1}{2} \left( \mathcal{L}_{hyb}(\hat{p}^\text{A}, \tilde{p}) + \mathcal{L}_{hyb}(\hat{p}^\text{B}, \tilde{p}) \right).
\end{equation}

\subsection{Dynamic Weight Adjustment}
\label{sec:GDWA}
To adaptively balance the joint optimization of $\mathcal{L}_{fusion}$ and $\mathcal{L}_{csc}$ in training Stage II, we introduce GDWA (Gradient-and-Descent-based Weight Adjustment), a dynamic weight adjustment strategy~based on two principles as described below. First, inspired by GDN \cite{chen2018gradnorm}, GDWA considers the gradient norm of each task, which reflects its impact on the shared fusion model parameters. Tasks with larger gradient norms are regarded as more significant and are assigned higher weights. Second, inspired by DWA \cite{liu2019end}, GDWA incorporates the descent rate, which is the ratio of the current loss to the previous loss, to measure convergence speed. Tasks with lower descent rates, indicating slower convergence and higher optimization difficulty, are given increased weights to ensure balanced optimization. For further discussion on GDWA, see Appendix \ref{ap.gdwa}.

Specifically, we define the gradient norm of task $k \in \{A, B\}$ as: $g_k = \sqrt{ \sum_{i} \| \nabla_{\theta_i} \mathcal{L}_k \|^2 }$, where $\theta_i$ denotes the shared parameters, and $\mathcal{L}_A = \mathcal{L}_{fusion}, \; \mathcal{L}_B = \mathcal{L}_{csc}$. Then, we normalize the gradient norms across tasks as: $\tilde{g}_A = g_A / (g_A + g_B), \quad \tilde{g}_B = 1 - \tilde{g}_A$. Additionally, to capture the relative convergence speed of each task, we compute the descent rate as: $r_k = \mathcal{L}_k^{j}/{\mathcal{L}_k^{j-1}}$, where $j$ denotes the the current epoch. We then convert the descent rates into weighting factors using a softmax function with a temperature parameter $T$: $s_A = e^{r_A / T}/(e^{r_A / T} + e^{r_B / T}), s_B = 1 - s_A$. Finally, the task weights are computed as a product of gradient norms and descent rates, followed by a normalization operation: $\lambda_A = (\tilde{g}_A s_A)/(\tilde{g}_A s_A + \tilde{g}_B s_B), \lambda_B = 1 - \lambda_A$. Therefore, in Eq. (\ref{equ:total_loss}), the dynamic weight of $\mathcal{L}_{csc}$ can be formulated as:
\begin{equation}
\label{eq.wcsc}
\omega_{csc} = \lambda_{csc} /\lambda_{fusion}.
\end{equation}

Overall, the proposed GDWA assigns greater importance to tasks with greater significance or slower convergence, thereby promoting balanced optimization across multiple tasks during training.

\section{Experiments and Results}
\label{sec:exp}
\begin{table*}[t]
\centering
\caption{Quantitative segmentation results on the FMB and MSRS datasets. Our unsupervised SSVIF achieves the best segmentation performance among unsupervised VIF methods and delivers competitive results compared to supervised approaches. Best and 2nd-best values are \textbf{highlighted} and \underline{underlined}.}
\resizebox{\textwidth}{!}{
\begin{tabular}{llcccccccccccccc}
\toprule
\multirow{2.5}{*}{Method} &  \multirow{2.5}{*}{Label} & 
\multicolumn{7}{c}{FMB Dataset (IoU \% ↑)}  &
\multicolumn{7}{c}{MSRS Dataset (IoU \% ↑)}  \\
\cmidrule(r){3-9} \cmidrule(r){10-16}
& & Back. & Per. & Road & Car & Bus & Lamp &mIoU & Back. & Car & Per. & Bike & Curve & Bump &mIoU \\
\midrule
CDDF.\cite{zhao2023cddfuse}  & w/o &36.14&70.82&\cellcolor{lightblue}\underline{91.52}&83.54&74.53&47.23&61.64&98.42 &90.48 &74.31  &68.94 & 50.95&74.21  &73.94\\
TIMF.\cite{liu2024task}  & w/o &32.69&65.39&89.77&83.35&65.86&43.42&59.58&98.37&90.54&71.10&67.24&51.04&75.53&73.95\\
SwinF.\cite{ma2022swinfusion}  & w/o &35.24&\cellcolor{lightred}\textbf{71.82}&91.41&83.44&73.85&47.75&61.25&98.35&90.46&72.31&66.73&48.38&73.65&72.78\\
EMMA \cite{zhao2024equivariant} & w/o &35.66&70.35&91.23&83.56&\cellcolor{lightblue}\underline{75.06}&47.29&61.46&98.42&90.50&74.28&69.10&51.15&72.46&74.06\\\midrule%\hdashline
MRFS \cite{zhang2024mrfs} & w  &36.10&70.48&90.86&83.47&\cellcolor{lightred}\textbf{75.24}&45.28&61.47&98.31&89.82&70.91&67.83&51.00&74.00&73.23\\
SAGE\cite{wu2025every}  & w  &36.89&70.75&90.86&83.65&74.57&\cellcolor{lightblue}\underline{48.10}&62.04&98.28&89.74&70.79&66.49&46.76&73.48&72.13\\
SeAF.\cite{tang2022image}  & w &\cellcolor{lightblue}\underline{37.53}&71.65&91.39&83.45&73.62&\cellcolor{lightred}\textbf{48.14}&61.77&\cellcolor{lightred}\textbf{98.55}&\cellcolor{lightred}\textbf{91.28}&\cellcolor{lightblue}\underline{75.18}&\cellcolor{lightred}\textbf{70.39}&\cellcolor{lightblue}\underline{56.09}&\cellcolor{lightred}\textbf{78.90}&\cellcolor{lightred}\textbf{76.41}\\
SegMiF \cite{liu2023segmif} & w &\cellcolor{lightred}\textbf{37.98}&66.81&\cellcolor{lightred}\textbf{91.59}&\cellcolor{lightred}\textbf{84.15}&74.74&44.82&\cellcolor{lightred}\textbf{62.14}&98.46&90.43&74.22&69.65&55.98&75.60&75.65\\\midrule
Ours  & w/o&36.68&\cellcolor{lightblue}\underline{71.71}&91.46&\cellcolor{lightblue}\underline{83.80}&73.73&47.32&\cellcolor{lightblue}\underline{62.07} &\cellcolor{lightblue}\underline{98.54}&\cellcolor{lightblue}\underline{91.23}&\cellcolor{lightred}\textbf{75.45}&\cellcolor{lightblue}\underline{70.11}&\cellcolor{lightred}\textbf{56.59}&\cellcolor{lightblue}\underline{78.17}&\cellcolor{lightblue}\underline{76.17}\\
\bottomrule
\end{tabular}}
\label{tab:seg_combined}
\end{table*}

\begin{figure*}[t]
	\centering
	\includegraphics[width=1\textwidth]{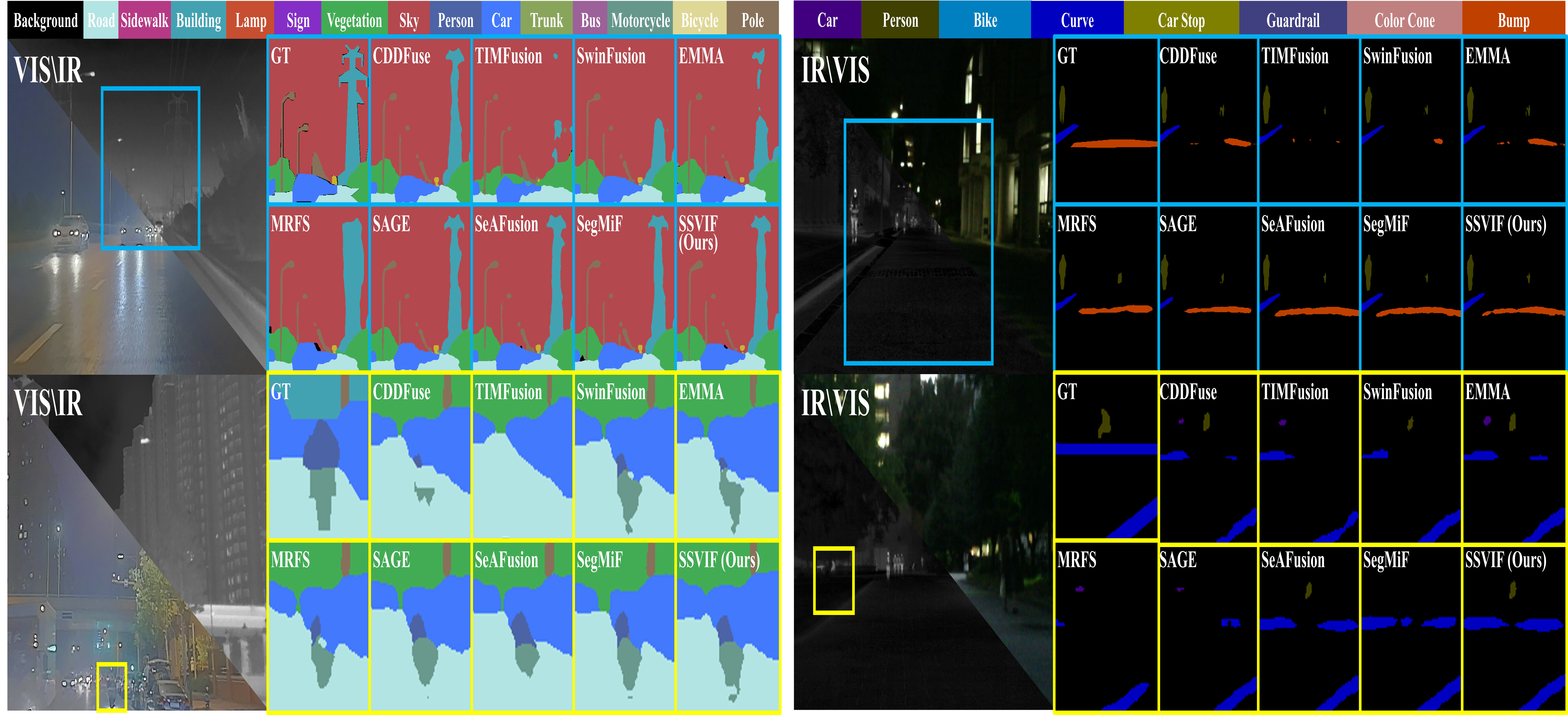}
	\caption{Qualitative segmentation results on the FMB (left) and MSRS (right) datasets. The fused images produced by our SSVIF (bottom-right corner) enable more accurate downstream segmentation, yielding clearer predictions for objects like buildings (left), pedestrians (both), and road curves (right).}
	\label{fig:qualitative_seg}
\end{figure*}

\subsection{Datasets and Implementation Details}
\label{sec:implement}
\noindent\textbf{Datasets.}~Two representative multi-modality datasets, namely FMB \cite{liu2023segmif} and MSRS \cite{tang2022piafusion}, are utilized for both training and evaluation. The FMB dataset comprises 1,500 infrared-visible image pairs annotated with 15 pixel-level semantic classes, while the MSRS dataset includes 1,444 image pairs with annotations for 9 classes. The image resolutions are $600 \times 800$ for FMB and $480 \times 640$ for MSRS. For training, 1,220 pairs from FMB and 1,083 from MSRS are used, with the remaining 280 (FMB) and 361 (MSRS) pairs reserved for evaluation.~The dataset split follows the original protocol defined in the corresponding paper.~It is worth noting that the segmentation labels are used solely for evaluation and are not involved in the training process of our SSVIF.

\noindent\textbf{Model Setup.}
As introduced in Section \ref{sec:SSVIF}, our SSVIF is a general training framework for fusion models. It can theoretically be applied to the training of any fusion model with a backbone-decoder architecture. In the evaluations of semantic segmentation and image fusion, our models are trained based on %the 
SwinFusion \cite{ma2022swinfusion} %fusion model 
using the FMB and MSRS training sets. To demonstrate the generalizability and flexibility of SSVIF, we further train the SeAFusion \cite{tang2022image} and EMMA \cite{zhao2024equivariant} fusion models using SSVIF in the ablation studies. For the segmentation model within the SSVIF training framework, we adopt SegFormer-B3 \cite{xie2021segformer}. All our models are trained from scratch using only the original network architectures, without relying on their original training frameworks or any pre-trained weights.

\noindent\textbf{Implementation Details.}~Before training, we split the original training sets of FMB and MSRS into training and validation subsets using a 9:1 ratio. All our models were trained for 110 epochs with early stopping (patience set to 10 epochs) to prevent overfitting. We used the Adam optimizer with an initial learning rate of $1\times10^{-4}$, a crop size of $160\times160$, and a batch size of 10. All our experiments were conducted using PyTorch on NVIDIA A100 and Tesla T4 GPUs.

\subsection{Compared Methods and Metrics}
\noindent\textbf{Compared Methods.}~We compare our SSVIF with eight state-of-the-art VIF methods including TIMFusion \cite{liu2024task} (TPAMI '24),  EMMA \cite{zhao2024equivariant} (CVPR '24), CDDFuse \cite{zhao2023cddfuse} (CVPR '23), SwinFusion \cite{ma2022swinfusion} (JAS '22), SAGE \cite{wu2025every} (CVPR '25), MRFS \cite{zhang2024mrfs} (CVPR '24),  SeAFusion \cite{tang2022image} (InfFus '22), and SegMiF \cite{liu2023segmif} (ICCV '23).~The first four are traditional VIF methods that do not require segmentation labels, while the latter four are application-oriented and rely on segmentation supervision.~For all compared methods, we use the default models provided in their publicly available code repositories.~All subsequent experiments are conducted to evaluate the performance of fused images themselves.

\noindent\textbf{Evaluation Metrics.}~To evaluate segmentation performance on fused images, we use the Intersection-over-Union (IoU) of each classes and mean IoU (mIoU). To evaluate fusion performance of fused images, we use seven metrics covering five different aspects of fused image quality, including two information theory-based metrics: EN and MI; one human perception inspired fusion metric: VIF; one image feature-based metric: $Q_{abf}$; two image structural similarity-based metrics: SSIM and MS-SSIM (MSS); and one color fidelity-based metric: $\Delta E$.~Except for $\Delta E$, higher values for all other metrics indicate superior fusion quality.~Metric calculations are performed following \cite{ma2019infrared,zhang2020vifb,gaurav2005ciede2000}.

\subsection{Semantic Segmentation Performance}
For fair comparison, we finetune the segmentation model provided by SegMiF \cite{liu2023segmif} using fused images generated by nine different fusion methods.~The backbone is Segformer-B3 \cite{xie2021segformer}.~We present quantitative segmentation results in Table \ref{tab:seg_combined}, where our SSVIF achieves the highest mIoU among fusion models trained without segmentation labels on both datasets. Compared to models trained with segmentation labels, SSVIF also delivers very competitive performance, further demonstrating its ability to guide fusion models in learning rich semantic information without label supervision. In addition, qualitative results in Fig. \ref{fig:qualitative_seg} further highlight the accurate segmentation results of SSVIF on both datasets. For more details on segmentation model training setup, see Appendix \ref{ap.seg_setup}.

\begin{table*}[t]
\centering
\caption{Quantitative fusion results on the FMB and MSRS datasets. Across both benchmarks, our unsupervised SSVIF consistently ranks among the top performers, achieving best or second-best results on most fusion metrics. Best and 2nd-best values are \textbf{highlighted} and \underline{underlined}.}
\resizebox{\textwidth}{!}{
\begin{tabular}{llcccccccccccccc}
\toprule
\multirow{2.5}{*}{Method} &  \multirow{2.5}{*}{Label} & 
\multicolumn{7}{c}{FMB Dataset (all ↑, except for $\Delta E$ ↓)}  &
\multicolumn{7}{c}{MSRS Dataset (all ↑, except for $\Delta E$ ↓)}  \\
\cmidrule(r){3-9} \cmidrule(r){10-16}
& &EN&MI  &VIF &$Q_{abf}$&SSIM&MSS&$\Delta E$&EN&MI  &VIF &$Q_{abf}$&SSIM&MSS&$\Delta E$ \\
\midrule
CDDF.\cite{zhao2023cddfuse}  & w/o & \cellcolor{lightblue}\underline{6.78} & \cellcolor{lightblue}\underline{4.15}  & \cellcolor{lightblue}\underline{0.87} & \cellcolor{lightblue}\underline{0.67} & 
\cellcolor{lightblue}\underline{1.00} & \cellcolor{lightblue}\underline{1.06} &6.11& 6.70 & \cellcolor{lightred}\textbf{4.71}  & \cellcolor{lightred}\textbf{1.03} & \cellcolor{lightred}\textbf{0.68} & 0.98 & \cellcolor{lightblue}\underline{1.02} &3.08\\
TIMF. \cite{liu2024task} & w/o & 6.49 & 3.12  & 0.56 & 0.54 & 0.81 & 0.77 &\cellcolor{lightblue}\underline{4.65}& \cellcolor{lightred}\textbf{6.92} & 2.77  & 0.63 & 0.43 & 0.56 & 0.73 &16.3\\
SwinF.\cite{ma2022swinfusion}  & w/o  & 6.53 & 3.85  & 0.77 & 0.65 & 0.96 & 1.00 &6.22& 6.43 & 3.69  & 0.85 & 0.60 & 0.89 & 0.99 &4.99\\
EMMA\cite{zhao2024equivariant}  & w/o  & 6.77 & 3.95  & 0.83 & 0.64& 0.90 & 1.03  &5.50& \cellcolor{lightblue}\underline{6.71} & 4.13  & 0.96 & 0.63 &0.94  & 1.03 &3.14\\\midrule
MRFS \cite{zhang2024mrfs} & w   & \cellcolor{lightblue}\underline{6.78} & 3.46  & 0.76 & 0.62 & 0.92 & 0.98 &7.16& 6.51 & 2.43   & 0.65 & 0.47 &0.75  & 0.89 &7.30\\
SAGE \cite{wu2025every} & w   &\cellcolor{lightred}\textbf{6.83} &3.43 &0.77 &0.64 &0.98 &\cellcolor{lightred}\textbf{1.10} & 8.04&6.00 &3.22 &0.71 &0.54 &0.89 &0.97 &5.24 \\
SeAF. \cite{tang2022image} & w  & 6.75 & 3.88  & 0.80 & 0.65 & 0.97 & 1.08 &6.17& 6.65 & 4.03  & 0.97 & \cellcolor{lightblue}\underline{0.67} & \cellcolor{lightblue}\underline{0.99}&\cellcolor{lightred}\textbf{1.05} &\cellcolor{lightred}\textbf{2.48}\\
SegMiF \cite{liu2023segmif} & w  & 6.51 & 3.01  & 0.61 & 0.43 & 0.91 & 1.04 &14.9& 6.08 & 2.22  & 0.62 & 0.37 & 0.80 & 0.97 &10.2\\\midrule
Ours  & w/o & 6.64 & \cellcolor{lightred}\textbf{4.76}  & \cellcolor{lightred}\textbf{0.88} & \cellcolor{lightred}\textbf{0.70} & \cellcolor{lightred}\textbf{1.02} & \cellcolor{lightblue}\underline{1.06} &\cellcolor{lightred}\textbf{4.34}&6.65 &\cellcolor{lightblue}\underline{4.33}    &\cellcolor{lightblue}\underline{1.02}  &\cellcolor{lightblue}\underline{0.67}  &\cellcolor{lightred}\textbf{1.03}  &\cellcolor{lightred}\textbf{1.05}  &\cellcolor{lightblue}\underline{2.57}\\
\bottomrule
\end{tabular}}
\label{tab:fus_combined}
\end{table*}
\begin{figure*}[ht]
	\centering
	\includegraphics[width=1\textwidth]{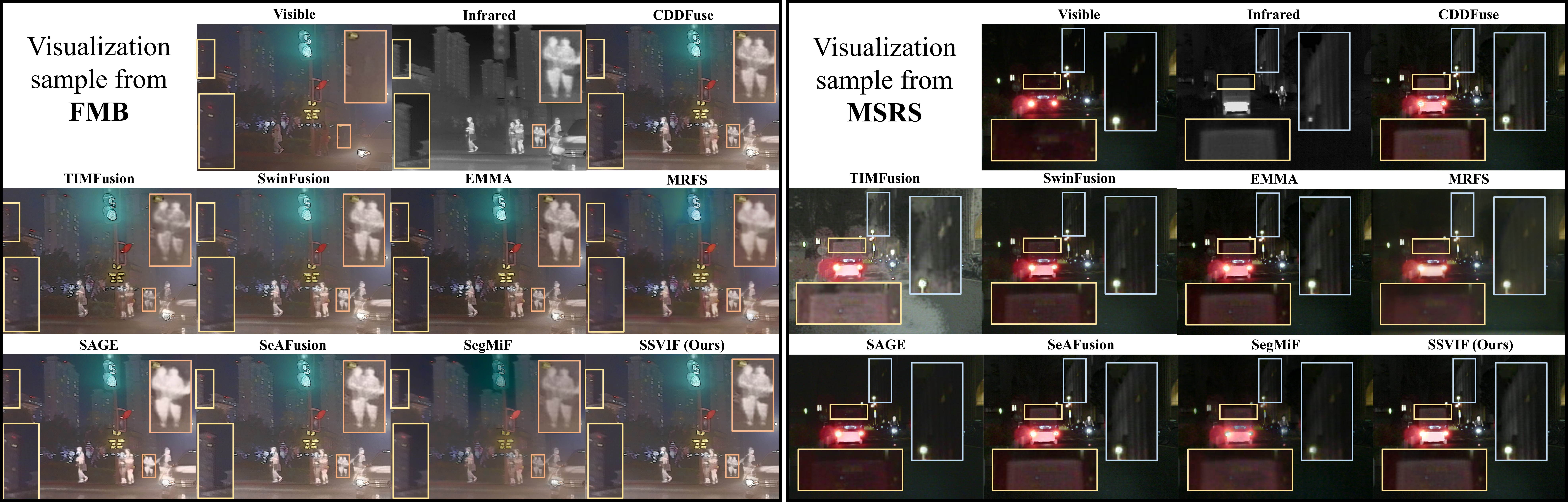}
	\caption{Qualitative fusion results on the FMB (left) and MSRS (right) datasets. Compared with existing methods, our SSVIF (bottom-right) produces fused images that better preserve both infrared saliency (e.g., pedestrians and vehicles) and visible structural details (e.g., traffic lights and building edges).}
	\label{fig:qualitative_fusion}
\end{figure*}

\subsection{Image Fusion Performance}
We present the quantitative fusion results on the FMB and MSRS datasets in Table \ref{tab:fus_combined}. SSVIF achieves the best or second-best performance across most metrics, including MI, VIF, $Q_{abf}$, SSIM, MSS, and $\Delta E$ on both datasets. Notably, it ranks first in SSIM on both datasets, highlighting its consistent ability to preserve structural details and produce visually coherent fusion results.~Visual results in Fig. \ref{fig:qualitative_fusion} further highlight the clear object boundaries and enhanced saliency achieved by SSVIF.

\subsection{Ablation Studies}
\label{sec:ablation}
To validate the rationality of our SSVIF, we conduct ablation studies on the FMB test set. Metrics including MI, $Q_{abf}$, SSIM, $\Delta E$, and mIoU are used to quantitatively evaluate both fusion and segmentation performance. The results of the experimental groups are summarized in Table \ref{table.Aba}.

\begin{table}[ht]
\centering
\caption{Ablation results on the FMB test set, confirming the effectiveness of each component. Best values are \textbf{highlighted}.}
\small  % 或者 \small，看版面空间
\setlength{\tabcolsep}{3.5pt}  % 缩小列间距
\begin{tabular}{cccccccc}
\toprule
Exp. & Model &Framework&MI &$Q_{abf}$&SSIM& $\Delta E$&mIoU\\
\cmidrule(r){2-8}
\multirow{7}{*}{\romannum{1}} &\multirow{2}{*}{SwinF.} & Original&3.85 & 0.65 & 0.96&6.22 &61.3   \\  
 & & SSVIF & \cellcolor{lightred}\textbf{4.76} & \cellcolor{lightred}\textbf{0.70} & 
\cellcolor{lightred}\textbf{1.02}&\cellcolor{lightred}\textbf{4.34} & \cellcolor{lightred}\textbf{62.1}   \\ \cmidrule(r){2-8} 
 &\multirow{2}{*}{EMMA}&Original&3.95 & 0.64 & 0.90&5.50 & 61.5   \\  
& & SSVIF & \cellcolor{lightred}\textbf{4.33} & \cellcolor{lightred}\textbf{0.67} & \cellcolor{lightred}\textbf{0.98}&\cellcolor{lightred}\textbf{4.36} & \cellcolor{lightred}\textbf{61.9}   \\ \cmidrule(r){2-8}  
 &\multirow{2}{*}{SeAF.}&Original&3.88 & 0.65 & \cellcolor{lightred}\textbf{0.97}&6.17 &61.8    \\  
 & & SSVIF & \cellcolor{lightred}\textbf{4.21} & \cellcolor{lightred}\textbf{0.67} & 0.96 &\cellcolor{lightred}\textbf{4.42}& \cellcolor{lightred}\textbf{62.0}   \\  \midrule  
Exp.&\multicolumn{2}{c}{Weight adjustment}&MI&$Q_{abf}$&SSIM& $\Delta E$&mIoU\\ \cmidrule(r){2-8}
\multirow{5}{*}{\romannum{2}} &\multicolumn{2}{c}{$w_{csc}=0.1$} & 4.73 & \cellcolor{lightred}\textbf{0.70} & 1.01&\cellcolor{lightred}\textbf{4.16} &61.7    \\
  &\multicolumn{2}{c}{UCW %\cite{kendall2018multi}
  }&4.43  & 0.68 & 1.01&4.68 &61.7   \\ 
 &\multicolumn{2}{c}{DWA %\cite{liu2019end}
 }& 4.69 & 0.69 & 1.01&4.29 &61.8   \\ 
 &\multicolumn{2}{c}{SegMiF %\cite{liu2023segmif}
 }&4.63   &0.69   &1.00  &4.22  &61.7   \\ %\cmidrule(r){2-8}
 &\multicolumn{2}{c}{GDWA (Ours)}& \cellcolor{lightred}\textbf{4.76} & \cellcolor{lightred}\textbf{0.70} & \cellcolor{lightred}\textbf{1.02}&4.34 & \cellcolor{lightred}\textbf{62.1}  \\
\midrule 
Exp. &\multicolumn{2}{c}{Configurations}&MI &$Q_{abf}$&SSIM& $\Delta E$&mIoU\\ \cmidrule(r){2-8}
\romannum{3}&\multicolumn{2}{c}{w/o $\mathcal{L}_{csc}$}&4.66   &0.69   &\cellcolor{lightred}\textbf{1.02}   &\cellcolor{lightred}\textbf{4.12}  &61.5    \\
\romannum{4}&\multicolumn{2}{c}{w/o 2-stage}& 4.72 & 0.69 &  1.01&4.25 & 61.7   \\
 &\multicolumn{2}{c}{w/ all (Ours)}& \cellcolor{lightred}\textbf{4.76} & \cellcolor{lightred}\textbf{0.70} & \cellcolor{lightred}\textbf{1.02}&4.34 & \cellcolor{lightred}\textbf{62.1}  \\
\bottomrule
\end{tabular}%}
\label{table.Aba}
% \vspace{-4mm} 
\end{table}

\noindent\textbf{SSVIF Training Framework.}
~In Exp.~\romannum{1}, to verify the effectiveness of proposed SSVIF training framework, we train three fusion models using SSVIF and compare them with the same models trained using their original training frameworks. The results demonstrate that the proposed SSVIF framework effectively provides additional semantic information during training, thereby improving the segmentation performance of the fused images. Moreover, fusion models trained with SSVIF are able to better preserve texture details and color distribution from the source images. Consistent improvements across different fusion models further demonstrate the generalizability of SSVIF, indicating that it can be applied to the training of various fusion model.

\noindent\textbf{Weight Adjustment Method}~In Exp.~\romannum{2}, to verify the effectiveness of the proposed GDWA, we train the same fusion model using different weight adjustment methods, including UCW \cite{kendall2018multi}, DWA \cite{liu2019end}, SegMiF \cite{liu2023segmif}, and a fixed setting where $w_{csc}$ in Eq. (\ref{equ:total_loss}) is set to 0.1. The results demonstrate that GDWA can effectively balance the fusion task and the cross-segmentation consistency task, allowing them to mutually reinforce each other, thereby improving the performance of the fusion model on both fusion and segmentation tasks. In particular, the results show that with GDWA, the fused images exhibit significantly better performance on the segmentation task, indicating that the method can more effectively inject useful semantic information into the fusion model during training.

\noindent\textbf{Cross-segmentation Consistency Loss $\mathcal{L}_{csc}$.}~Then, in Exp.~\romannum{3}, we eliminate the cross-segmentation consistency loss $\mathcal{L}_{csc}$ in Eq. (\ref{equ:total_loss}). The results demonstrate that, $\mathcal{L}_{csc}$ can effectively guide the integration of semantic information into the model training, thereby enhancing the segmentation performance on the fused images. It is worth noting that since the segmentation model and the segmentation head participate in training solely through $\mathcal{L}_{csc}$, they are also removed in Exp.~\romannum{3}~when this loss term is eliminated to avoid redundant computation.

\noindent\textbf{Two-stage Training.}~In Exp.~\romannum{4}, we abandon the 2-stage training and train the fusion model, segmentation model, and segmentation head simultaneously.
~The suboptimal results demonstrates that the 2-stage training strategy can effectively reduce training difficulty and improve training robustness.

In summary, the results in Table \ref{table.Aba} validate the effectiveness and rationality of our proposed methods.

\subsection{Gradient Analysis}
To verify the correlation between fusion and CSC tasks, we apply the gradient projection method to assess the influence of the proposed CSC task on the fusion task. After a complete training process of SSVIF, the average projection coefficient of CSC gradient onto fusion task is $1.05e^{-2}$. The average projection coefficient is positive, meaning that the CSC task~facilitates the optimization of the fusion model towards generating higher-quality fused images.~For detailed explanation of gradient projection and projection coefficient, see Appendix \ref{ap.gradient}.

\begin{table}[ht]
\centering
\caption{Computational efficiency during inference, which remains nearly unchanged when applying our SSVIF to existing fusion models.}
\small  % 或者 \footnotesize 如果需要再小一点的字体
\setlength{\tabcolsep}{2pt}  % 缩小列间距，默认值是6pt
\begin{tabular}{ccccc}
\toprule 
Model & Input size & Params (M) & FLOPs (G) & Mem (MB)\\
\midrule 
SwinF.\_Original & (256,256,1) & 0.97 & 75.98 & 683.4 \\  
SwinF.\_SSVIF    & (256,256,3) & 0.98 & 76.07 & 684.4 \\ 
\midrule 
EMMA\_Original   & (256,256,1) & 1.52 & 8.62  & 49.4 \\  
EMMA\_SSVIF      & (256,256,3) & 1.52 & 8.65  & 50.4 \\ 
\midrule 
SeAF.\_Original  & (256,256,1) & 0.17 & 10.88 & 105.2 \\  
SeAF.\_SSVIF     & (256,256,3) & 0.17 & 10.94 & 106.2 \\ 
\bottomrule 
\end{tabular}
\label{table.Compu}
\end{table}

\subsection{Computational Efficiency}
Table~\ref{table.Compu} shows that during inference, fusion models trained with SSVIF retain comparable model size and computational efficiency to their original versions, with only slight differences due to the use of three-channel input data.~This indicates that SSVIF can enhance the performance of fusion models while keeping their original architecture and parameter count almost unchanged.~It is worth noting that the dual segmentation branches (Section \ref{sec:SSVIF})~in the SSVIF training framework are used only during training and have no impact on inference.

\subsection{Semantic Verification with Object Detection}
Our CSC loss can help our model to learn more semantic features, which may help other downstream tasks as well in addition to semantic segmentation. To test this, we run object detection on the fused images.~Firstly, we train the detection models (YOLOv11 \cite{Jocher_Ultralytics_YOLO_2023}) by the fused images from various VIF methods.~Then we get quantitative detection performance of different VIF methods as shown in Table \ref{table.M3FD_detection_small}, where our SSVIF achieves the best performance among unsupervised VIF methods. Compared to  supervised application-oriented VIF methods, SSVIF also delivers competitive results. These results suggest that SSVIF effectively enhances the semantic content of fused images, highlighting its potential for broader application in high-level vision tasks beyond semantic segmentation. For details on detection training and additional results, see Appendix~\ref{ap:detect}.

\begin{table}[t]
\centering
\caption{Detection results on M$^3$FD \cite{liu2022target}, demonstrating our SSVIF learns useful semantic information that can generalize to non-segmentation tasks such as detection. Best and 2nd-best values are \textbf{highlighted} and \underline{underlined}.}
\small
\setlength{\tabcolsep}{6.5pt}
\begin{tabular}{cccccc}
\toprule
\multicolumn{6}{c}{\textbf{Comparison with Unsupervised Methods}} \\
\midrule
Method & CDDF. & TIMF. & SwinF. & EMMA & Ours \\
mAP@0.5 & 79.39 & 79.66 & \cellcolor{lightblue}\underline{80.03} & 79.50 & \cellcolor{lightred}\textbf{80.18} \\
\midrule
\multicolumn{6}{c}{\textbf{Comparison with Supervised Methods}} \\
\midrule
Method & MRFS & SAGE & SeAF. & SegMiF & Ours \\
mAP@0.5 & 78.92 & \cellcolor{lightred}\textbf{80.53} & 80.18 & \cellcolor{lightblue}\underline{80.41} & 80.18 \\
\bottomrule
\end{tabular}
\label{table.M3FD_detection_small}
\end{table}

\section{Conclusions}
In this paper, we propose SSVIF, a general self-supervised training framework for segmentation-oriented visible and infrared image fusion. Leveraging the consistency between feature-level fusion-based segmentation and pixel-level fusion-based segmentation, we introduce a novel self-supervised task, cross-segmentation consistency, that enables the fusion model to learn high-level semantic features without supervision of downstream task labels. To further improve performance of this framework, we design a two-stage training strategy and a dynamic weight adjustment method that effectively balance joint task optimization. Extensive experiments demonstrate the effectiveness of SSVIF and validate the design of the self-supervised task and other key components, providing a strong foundation for future research in VIF.

\section*{Acknowledgments}
This study has received funding from the Royal Society Research Grant (No. RG\textbackslash{}R1\textbackslash{}251462).

{\appendices
\section{More discussions}
\subsection{More Discussion on GDWA}
\label{ap.gdwa}
Our GDWA is mainly inspired by DWA \cite{liu2019end}, which considers descent rate of different task losses for joint training's balance between different tasks. Different from DWA, we further present the task preference to distinguish the primary and secondary tasks. Unlike SegMiF \cite{liu2023segmif}, which manually fixes the task preference factor, we are inspired by GDN \cite{chen2018gradnorm} and utilize the gradient norm of each task loss to automatically reflect the relative importance of different tasks. Compared to DWA and SegMiF, our GDWA can more effectively and automatically prioritize the training of primary tasks. In contrast to GDN, our GDWA eliminates the reliance on early loss, resulting in a more robust training process.

\subsection{More Details on Segmentation Model Training}
\label{ap.seg_setup}

\noindent\textbf{Setup.}~For fair comparison, we finetune the segmentation model provided by SegMiF \cite{liu2023segmif} using fused images generated by nine different fusion methods. The backbone is Segformer-B3 \cite{xie2021segformer}. All segmentation models are trained with cross-entropy loss and optimized using AdamW. The finetuning is performed over 80 epochs with a batch size of 12, where the backbone is frozen for the first 30 epochs. The initial learning rate is set to $5 \times 10^{-4}$ and reduced to $3 \times 10^{-5}$ after unfreezing the backbone, following a cosine annealing schedule. Early stopping with a patience of 10 epochs is employed to prevent overfitting. Since the test sets of FMB and MSRS contain different segmentation categories, two separate segmentation models are trained for each fusion method. The dataset splits during finetuning follow the protocols in \cite{liu2023segmif, tang2022piafusion}.

\subsection{Limitations and Broader Impacts}
\label{ap.limitations}
\noindent\textbf{Limitations.}~As a flexible training framework for visible and infrared fusion models, the performance of SSVIF significantly depends on the performance of the chosen fusion models. In the future, we will design a more effective fusion network to further improve fusion performance.

\noindent\textbf{Broader Impacts.}~This paper aims to broaden the applicability of multi-modal data to diverse downstream tasks and research domains. However, this broader scope may introduce challenges when applying the model in tasks or domains involving harmful content. These challenges arise from the data rather than the model itself. Therefore, it is essential to have adequate data regularization to mitigate potential risks and ensure responsible use of our model.

\section{More experimental results}
\subsection{More explanations of hyperparameter settings}
\label{ap.hyperparameter}
\noindent\textbf{The class number of dual segmentation branches: $n$.}~For the cross-segmentation consistency task, the class number of the segmentation head and the segmentation model should be the same. To determine the value of $n$ in our practical experiments, we refer to the class number settings in two widely used multimodal segmentation datasets: 15 for FMB \cite{liu2023segmif} and 9 for MSRS \cite{tang2022piafusion}. We compared the results of $n=15$ and $n=9$ in an ablation study shown in Table \ref{table.classnumber}. Through experiments, we found that setting $n=15$ yields better segmentation performance. Therefore, we adopt $n=15$ as the practical setting in our~experiments of SSVIF.
\begin{table}[htbp]
\caption{Ablation study for class number on the FMB dataset. As 15 yields better performance, we adopt this setting in our experiments. Best values are \textbf{highlighted}.}
\label{table.classnumber}
\centering
\small
\setlength{\tabcolsep}{3pt}
\begin{tabular}{ccccccccc}
\toprule
{Configurations} &EN&MI  &VIF &$Q_{abf}$&SSIM&MSS&$\Delta E$& mIoU \\
\midrule
 $n=9$ & \textbf{6.64} & 4.72  & \textbf{0.89} & \textbf{0.70} & 1.01 & \textbf{1.06} &\textbf{4.08}&61.76\\
$n=15$& \textbf{6.64} & \textbf{4.76}  & 0.88 & \textbf{0.70} & \textbf{1.02} & \textbf{1.06} &4.34&\textbf{62.07}\\
\bottomrule
\end{tabular}
\end{table}

\noindent\textbf{The hyperparameters in fusion loss (Eq. (\ref{eq.fusion})): $\lambda_1,\lambda_2,\lambda_3,\lambda_4$.}~As decried in Section \ref{sec:implement}, in the evaluations of semantic segmentation and image fusion, our SSVIF models are trained based on SwinFusion \cite{ma2022swinfusion}. In the previous research of SwinFusion model \cite{ma2022swinfusion}, the hyperparameter for intensity loss, gradient loss, and structural similarity loss are $\lambda_1=20$, $\lambda_2=20$, and $\lambda_3=10$, respectively. Therefore, we followed these settings in our practical experiments for SSVIF. For the color-preserving loss, which was not considered in most existing VIF methods, we did ablation studies to find the appropriate value for $\lambda_4$. As shown in Table \ref{table.colorloss}, $\lambda_4=20$ leads to better fusion and segmentation performance. Therefore, we adopt $\lambda_4=20$ in our practical experiments for SSVIF.
\begin{table}[htbp]
\caption{Ablation study for $\lambda_4$ on the FMB dataset. As 20 yields better performance, we adopt this setting in our experiments. Best values are \textbf{highlighted}.}
\label{table.colorloss}
\centering
\small
\setlength{\tabcolsep}{3pt}
\begin{tabular}{ccccccccc}
\toprule
Configurations &EN&MI  &VIF &$Q_{abf}$&SSIM&MSS&$\Delta E$& mIoU \\
\midrule
 $\lambda_4=10$ & 6.62 & 4.68  & 0.87 & 0.69 & 1.01 & 1.05 &4.45&61.88\\
$\lambda_4=20$& \textbf{6.64} & \textbf{4.76}  & \textbf{0.88} & \textbf{0.70} & \textbf{1.02} & \textbf{1.06} &4.34&\textbf{62.07}\\
$\lambda_4=30$& 6.61 & 4.61  & 0.87 & 0.69 & 1.01 & \textbf{1.06} &\textbf{4.17}&61.48\\
\bottomrule
\end{tabular}
\end{table}

\subsection{Three-channel Fusion v.s. Single-channel Fusion}
\begin{table}[h]
\caption{Ablation study for three-channel fusion on the FMB dataset, demonstrating its effectiveness compared to single-channel fusion. Best values are \textbf{highlighted}.}
\label{table.3chanel}
\centering
\small
\setlength{\tabcolsep}{3pt}
\begin{tabular}{ccccccccc}
\toprule
Configurations &EN&MI  &VIF &$Q_{abf}$&SSIM&MSS&$\Delta E$& mIoU \\
\midrule
 single-channel & 6.61 & 4.61  & \textbf{0.88} & 0.69 & 0.96 & 0.99 &\textbf{3.95}&61.16\\
three-channel& \textbf{6.64} & \textbf{4.76}  & \textbf{0.88} & \textbf{0.70} & \textbf{1.02} & \textbf{1.06} &4.34&\textbf{62.07}\\
\bottomrule
\end{tabular}
\end{table}
In SSVIF, unlike most existing VIF methods \cite{ma2022swinfusion, zhao2023cddfuse, zhao2024equivariant}, we do not use single-channel (grayscale) images as the input and output of the fusion model. Instead, we choose three-channel (RGB) images as input. This is because the color information is useful for our proposed cross-segmentation consistency task. To validate this hypophysis, we also did an additional ablation study on FMB dataset for three-channel and single-channel fusion, in which we modify the model’s input and output from three channels to a single channel. The results  %as shown 
in Table \ref{table.3chanel} indicate that three-channel fusion more effectively improves segmentation performance and visual quality by fully leveraging color information.

\subsection{More Object Detection Settings and Results}
\label{ap:detect}

\noindent\textbf{Setup.}~Specifically, we generate fused images using different fusion methods on the M$^3$FD dataset \cite{liu2022target}, which contains 4,200 pairs of infrared and visible images with detection annotations for six classes. The dataset is split into training, validation, and test sets in an 8:1:1 ratio. A YOLOv11n \cite{Jocher_Ultralytics_YOLO_2023} detector is then retrained on the fused images produced by each method for 100 epochs, using the default configuration settings of YOLOv11 \cite{Jocher_Ultralytics_YOLO_2023}. Detection performance is evaluated using AP@0.5 and mAP@0.5 metrics. Additional quantitative detection results are shown in Table \ref{table.M3FD_detection} and qualitative detection results are shown in Figs. \ref{fig:qualitative_detection} and \ref{fig:qualitative_detection_more}.

\begin{table*}[htbp]
\caption{Quantitative detection results on the M$^3$FD datasets. Our SSVIF achieves 80.18 mAP@0.5, performing on par with supervised methods and obtaining the highest AP for Truck, demonstrating its ability to learn transferable semantic features for detection. Best and 2nd-best values are \textbf{highlighted} and \underline{underlined}.}
\label{table.M3FD_detection}
\centering
\small
%\resizebox{\textwidth}{!}{
\begin{tabular}{lllccccccc}
\toprule
\multirow{2.5}{*}{\hspace{0em}Method} &\multirow{2.5}{*}{\hspace{0em}Source}  &\multirow{2.5}{*}{\hspace{0em}Label}&\multicolumn{7}{c}{Average Precision at IoU=0.5 (AP@0.5 \%↑)} \\\cmidrule{4-10}
 & & &Per.&Car &Bus &Mot. &Lamp&Truck&mAP@0.5\\
\midrule
CDDF.\cite{zhao2023cddfuse}&CVPR'23&w/o &77.93   &    89.84  &      91.90     &  56.07   &    71.88     &   88.70& 79.39\\  
TIMF.\cite{liu2024task}&TPAMI'24&w/o & 77.89  &     89.93  &     89.78    &     \cellcolor{lightred}\textbf{59.00}&       \cellcolor{lightblue}\underline{74.23}  &     87.11 &79.66\\
SwinF.\cite{ma2022swinfusion}&JAS'22&w/o &79.53  &     90.45    &    91.30  &     55.99  &     73.24    &   \cellcolor{lightblue}\underline{89.65}& 80.03\\
EMMA\cite{zhao2024equivariant}&CVPR'24&w/o &78.42   &    90.33   &     91.80   &    56.57  &     72.06    &    87.80& 79.50\\ \midrule 
MRFS\cite{zhang2024mrfs}&CVPR'24&w/ &77.64   &    89.99   &    91.85    &      55.00   &    70.07  &     88.97 &78.92\\
  SAGE\cite{wu2025every}&CVPR'25&w/ &79.22   &    \cellcolor{lightblue}\underline{90.48}  &     91.66   &    57.55  &     \cellcolor{lightred}\textbf{76.03}     &  88.27& \cellcolor{lightred}\textbf{80.53}\\
 SeAF.\cite{tang2022image}&InfFus'22&w/ &79.71   &    90.42  &     \cellcolor{lightred}\textbf{92.81}   &   \cellcolor{lightblue}\underline{58.99}  &     70.93   &    88.26& 80.18\\
 SegMiF\cite{liu2023segmif}&ICCV'23&w/ &\cellcolor{lightred}\textbf{80.04}  &     \cellcolor{lightred}\textbf{91.06}  &     91.58   &    56.94   &    74.07     &  88.77& \cellcolor{lightblue}\underline{80.41}\\ \midrule 

Ours&Proposed&w/o &\cellcolor{lightblue}\underline{79.82}  &     90.12   &    \cellcolor{lightblue}\underline{92.62}   &    56.02   &    72.84    &   \cellcolor{lightred}\textbf{89.67}& 80.18\\
\bottomrule
\end{tabular}%}
\end{table*}

\begin{figure*}
	\centering
	\includegraphics[width=1\textwidth]{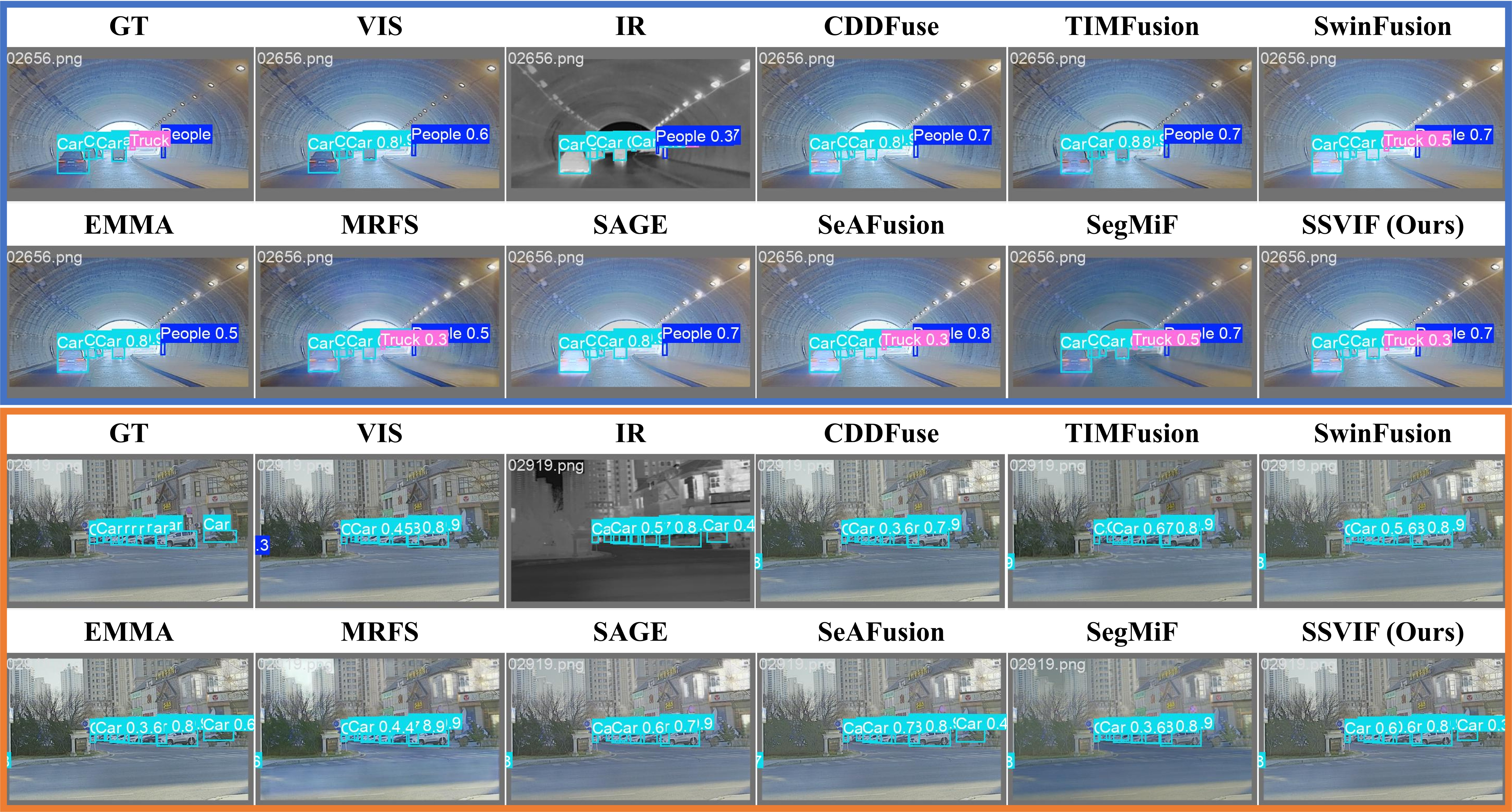}
	\caption{Qualitative detection results on the M$^3$FD dataset. Compared with other fusion methods, our SSVIF (bottom-right) generates fused images that provide more reliable cues for object detection, leading to clearer recognition of multiple categories (e.g., people, cars, and trucks). Notably, SSVIF yields more stable bounding boxes with higher confidence scores, while alternative approaches often miss small objects or produce uncertain predictions. These results demonstrate the effectiveness of SSVIF in enhancing downstream detection performance through semantically enriched fusion.}
    \label{fig:qualitative_detection_more}
\end{figure*}

\begin{figure*}
	\centering
	\includegraphics[width=1\textwidth]{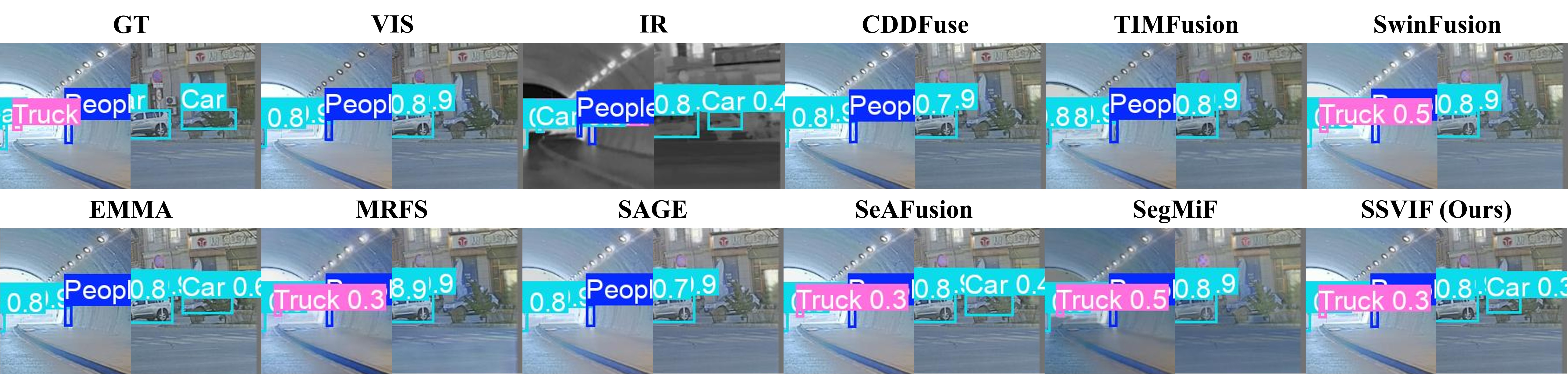}
	\caption{More detailed qualitative detection results on the M$^3$FD dataset.}
	\label{fig:qualitative_detection}
\end{figure*}

\section{Detailed Explanation of Gradient Projection and Projection Coefficient}
\label{ap.gradient}
\begin{figure}[htbp]
	\centering
	\includegraphics[width=0.45\textwidth]{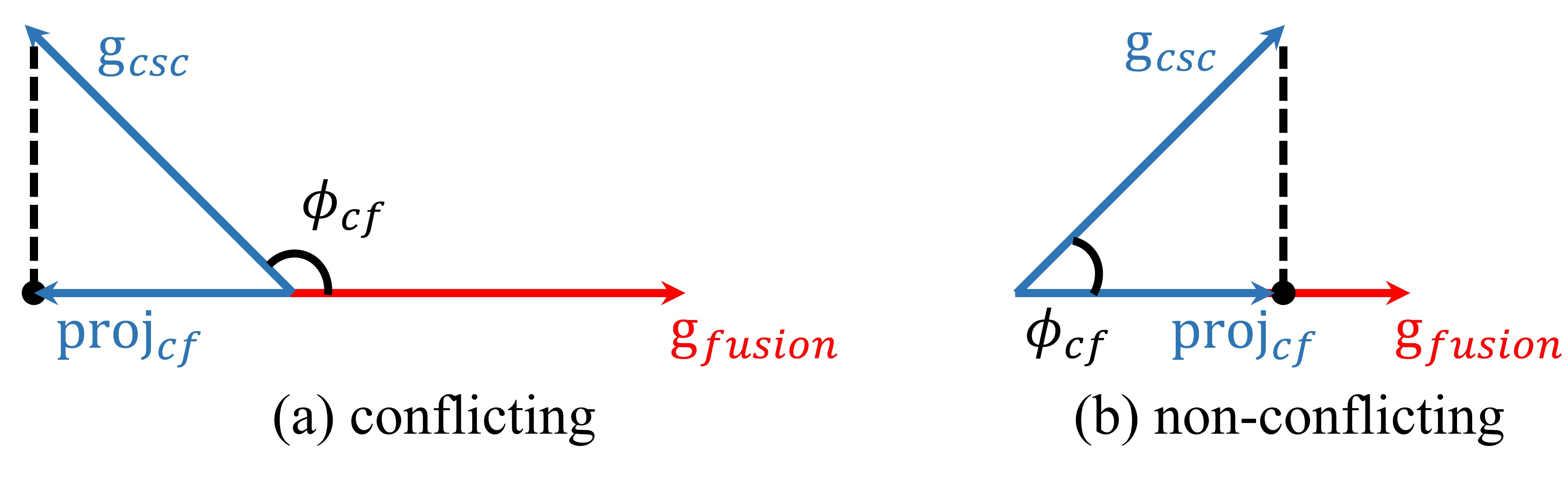}
	\caption{Visualization of conflicting vs. non-conflicting gradients.}
    \label{fig:gradient}
\end{figure}

For SSVIF's training process, the goal of joint training with fusion and cross-segmentation consistency (CSC) task is to find parameters $\theta$ of a fusion model $f_\theta$ that achieve high average performance across both tasks. More formally, we aim to solve the problem: $\min_\theta\mathbb{E}_{i\sim\{fusion, csc\}}[\mathcal{L}_{i}(\theta)]$. We denote the gradient of each task as $\mathbf{g}_i = \nabla \mathcal{L}_i(\theta)$, where $i\sim\{fusion, csc\}$. (We drop the reliance on $\theta$ in the notation for brevity.) Based on previous research \cite{yu2020gradient}, we can have the condition as below.

\noindent\textbf{Definition 1.}~\textit{
We define $\phi_{cf}$ as the angle between two task gradients $\mathbf{g}_{csc}$ and $\mathbf{g}_{fusion}$. We define the gradients as \textbf{conflicting} when $\cos\phi_{cf} < 0$.}

\noindent\textbf{Definition 2.}~\textit{
We define the \textbf{projection coefficient} of the CSC gradient $\mathbf{g}_{csc}$ onto the fusion gradient $\mathbf{g}_{fusion}$ as}
\begin{equation}
\label{equ:proj}
\alpha_{cf} = \frac{\mathbf{g}_{csc} \cdot \mathbf{g}_{fusion}}{\|\mathbf{g}_{fusion}\|^2}.
\end{equation}
\textit{The \textbf{projection vector} is then defined as} $
\text{proj}_{cf} = \alpha_{cf} \, \mathbf{g}_{fusion}$\textit{, which represents the component of $\mathbf{g}_{csc}$ in the direction of $\mathbf{g}_{fusion}$.
}

\noindent\textbf{Definition 3.}~\textit{
We define the \textbf{cosine similarity} between the CSC gradient $\mathbf{g}_{csc}$ and the fusion gradient $\mathbf{g}_{fusion}$ as} 
\begin{equation}
\label{equ:cos}
\cos\phi_{cf} = \frac{\mathbf{g}_{csc} \cdot \mathbf{g}_{fusion}}{\|\mathbf{g}_{csc}\| \cdot \|\mathbf{g}_{fusion}\|}.
\end{equation}

Therefore, combining Eq.~(\ref{equ:proj}) and Eq.~(\ref{equ:cos}), we have
\[
\alpha_{cf} = \frac{\|\mathbf{g}_{csc}\|}{\|\mathbf{g}_{fusion}\|} \cos\phi_{cf}.
\]
According to Definition~1, we derive the following corollary.

\noindent\textbf{Corollary 1.}~\textit{
If the projection coefficient $\alpha_{cf} < 0$, then the CSC gradient $\mathbf{g}_{csc}$ and the fusion gradient $\mathbf{g}_{fusion}$ are conflicting.
}

Consequently, a positive projection coefficient (i.e., $\alpha_{cf} > 0$) indicates that $\mathbf{g}_{csc}$ and $\mathbf{g}_{fusion}$ are non-conflicting (Fig. \ref{fig:gradient}). In this case, the CSC gradient has a beneficial effect on the optimization of the fusion task during joint training. Symmetrically, a positive projection coefficient of $\mathbf{g}_{fusion}$ onto $\mathbf{g}_{csc}$ implies that the fusion task provides a beneficial gradient signal for the CSC objective.

}

{%\small
\bibliographystyle{IEEEtran}
\bibliography{main}
}

 \begin{IEEEbiography}[{\includegraphics[width=1in,height=1.25in]{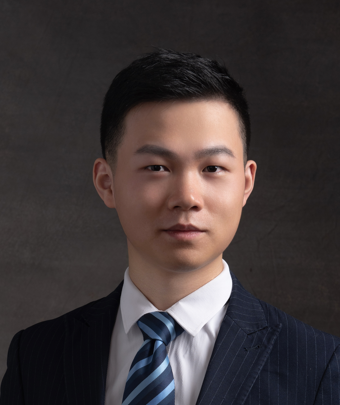}}]{Zixian Zhao } received the B.Sc. degree in automation from the School of Mechanical Engineering and Automation, Harbin Institute of Technology (Shenzhen) in 2022 and an MSc in Applied Machine Learning degree with distinction from the Department of Electrical and Electronic Engineering, Imperial College London in 2023. He is currently a PhD student at the Fusion Intelligence Laboratory in the Department of Computer Science at the University of Exeter.
\end{IEEEbiography}

\begin{IEEEbiography}[{\includegraphics[width=1in,height=1.25in]{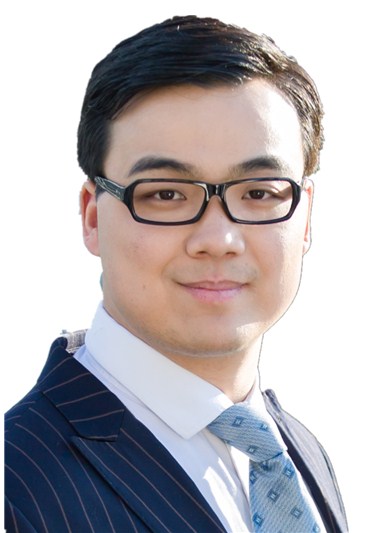}}]{Xingchen Zhang (M'21)} received his B.Sc. degree from Huazhong University of Science and Technology in 2012, and his Ph.D. degree from Queen Mary University of London in 2018. He is currently a Senior Lecturer and the Director of the Fusion Intelligence Laboratory in the Department of Computer Science at the University of Exeter. Previously, he was a Visiting Researcher and Marie Skłodowska-Curie Individual Fellow at the Personal Robotics Laboratory, Department of Electrical and Electronic Engineering, Imperial College London. He also previously worked as a Teaching Fellow and Research Associate in the same department at Imperial College London. His main research interests include image fusion, human motion and intention prediction, multimodal robot vision, and object tracking. He has been listed among the World’s Top 2\% Scientists from 2023 to 2025 (Stanford University’s list). He is %a member of the UKRI Talent Peer Review College and 
a Fellow of the Higher Education Academy.
\end{IEEEbiography}

\vfill

\end{document}